\documentclass[11pt,a4paper]{article}
\usepackage[hyperref]{AACL-IJCNLP2020}
\usepackage{times}
\usepackage{latexsym}
\usepackage{graphicx}
\usepackage{multirow}
\usepackage{array}
\usepackage{booktabs}
\usepackage{amsmath}
\usepackage{amssymb}
\usepackage{float}
\usepackage{url}
\usepackage{xcolor}

\usepackage[toc,page]{appendix}

\newcommand{\printfnsymbol}[1]{%
  \textsuperscript{\@fnsymbol{#1}}%
}

\usepackage{microtype}

\aclfinalcopy 
\def\aclpaperid{95} 

\title{Neural RST-based Evaluation of Discourse Coherence}

\author{
Grigorii Guz\thanks{\hspace{0.02in} Authors contributed equally} \hspace{0.02in}$^{1}$, Peyman Bateni$^*$$^{1,2}$, Darius Muglich$^1$, Giuseppe Carenini$^1$ \\
University of British Columbia$^1$, Inverted AI$^2$\\
\texttt{\{g.guz@cs, pbateni@cs, darius.muglich@alumni, carenini@cs\}.ubc.ca}
}

\date{}

\begin{document}
\maketitle
\begin{abstract}
This paper evaluates the utility of Rhetorical Structure Theory (RST) trees and relations in discourse coherence evaluation. We show that incorporating silver-standard RST features can increase accuracy when classifying coherence. We demonstrate this through our tree-recursive neural model, namely RST-Recursive, which takes advantage of the text's RST features produced by a state of the art RST parser. We evaluate our approach on the Grammarly Corpus for Discourse Coherence (GCDC) and show that when ensembled with the current state of the art, we can achieve the new state of the art accuracy on this benchmark. Furthermore, when deployed alone, RST-Recursive achieves competitive accuracy while having 62\% fewer parameters.

\end{abstract}

\section{Introduction}

Discourse coherence has been the subject of much research in Computational Linguistics thanks to its widespread applications \cite{lai-grammerly}. Most current methods can be described as either stemming from explicit representations based on the Centering Theory \cite{grosz-centering-theory}, or deep learning approaches that learn without the use of hand-crafted linguistic features.

Our work explores a third research avenue based on the Rhetorical Structure Theory (RST) \cite{mann-RST-theory}. We hypothesize that texts of low/high coherence tend to adhere to different discourse structures. Thus, we pose that using even silver-standard RST features should help in separating coherent texts from incoherent ones. This stems from the definition of the coherence itself - as the writer of a document needs to follow specific rules for building a clear narrative or argument structure in which the role of each constituent of the document should be appropriate with respect to its local and global context, and even existing discourse parsers should be able to predict a plausible structure that is consistent across all coherent documents. However, if a parser has difficulty interpreting a given document, it will be more likely to produce unrealistic trees with improbable patterns of discourse relations between constituents. This idea was first explored by \citeauthor{feng-etal-2014-impact} \shortcite{feng-etal-2014-impact}, who followed an approach similar to \citeauthor{Barzilay-Entity-Grid} \shortcite{Barzilay-Entity-Grid} by estimating entity transition likelihoods, but instead using discourse relations (predicted by a state of the art discourse parser \cite{feng-hirst-2014-linear}) that entities participate in as opposed to their grammatical roles. Their method achieved significant improvements in performance even when using silver-standard discourse trees, showing potential in the use of parsed RST features for classifying textual coherence. 

\begin{figure}[t]
    \centering
    \includegraphics[width=3.1in]{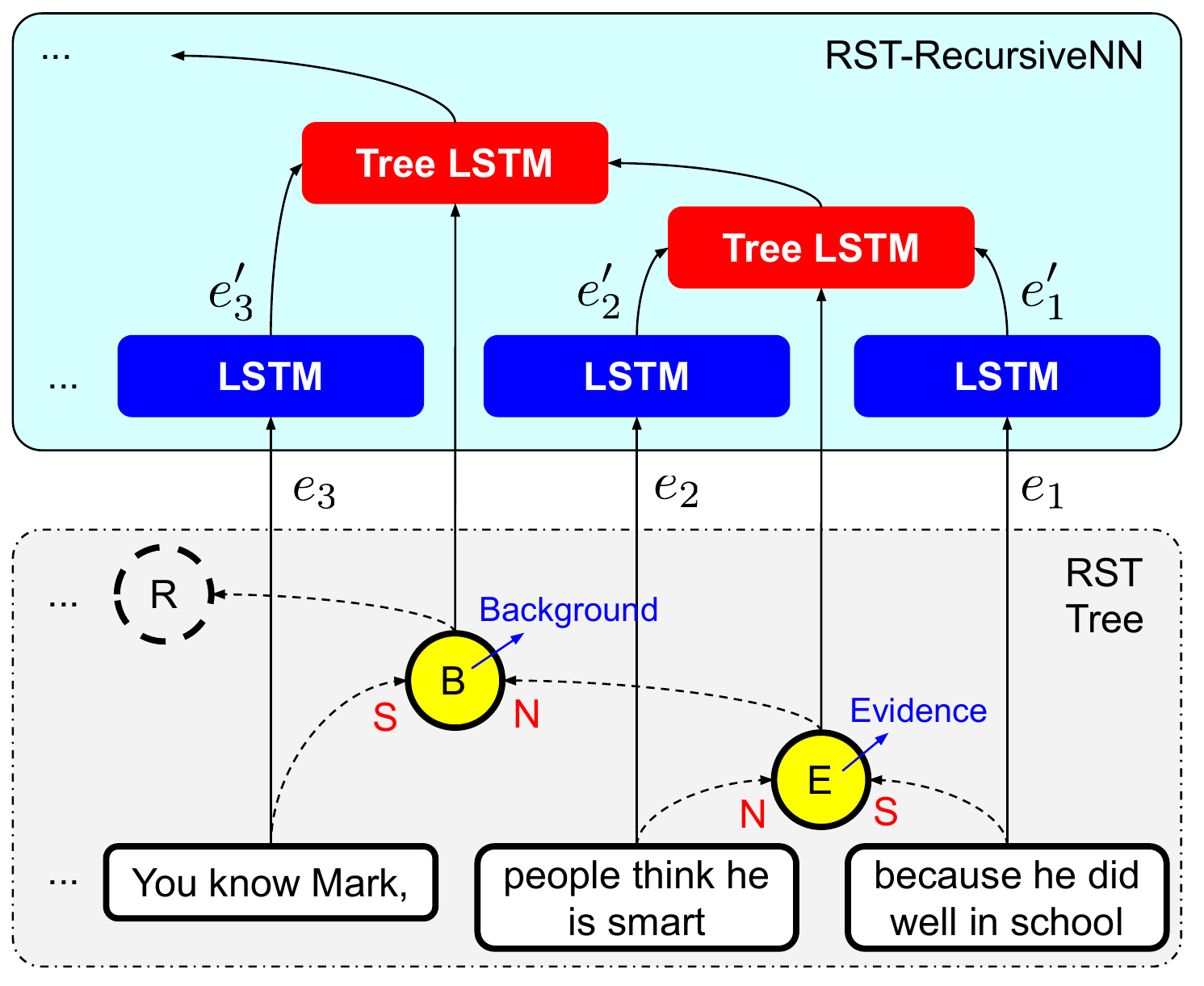}
    \vspace{-0.3in}
    \caption{Overview of RST-Recursive; EDU embeddings are generated for the leaf nodes using the EDU network. Subsequently, the RST tree is recursively traversed bottom-up using the RST network.}
    \label{fig:rst-recursive-overview}
    \vspace{-0.25in}
\end{figure}

\begin{figure}
    \centering
    \includegraphics[width=3.05in]{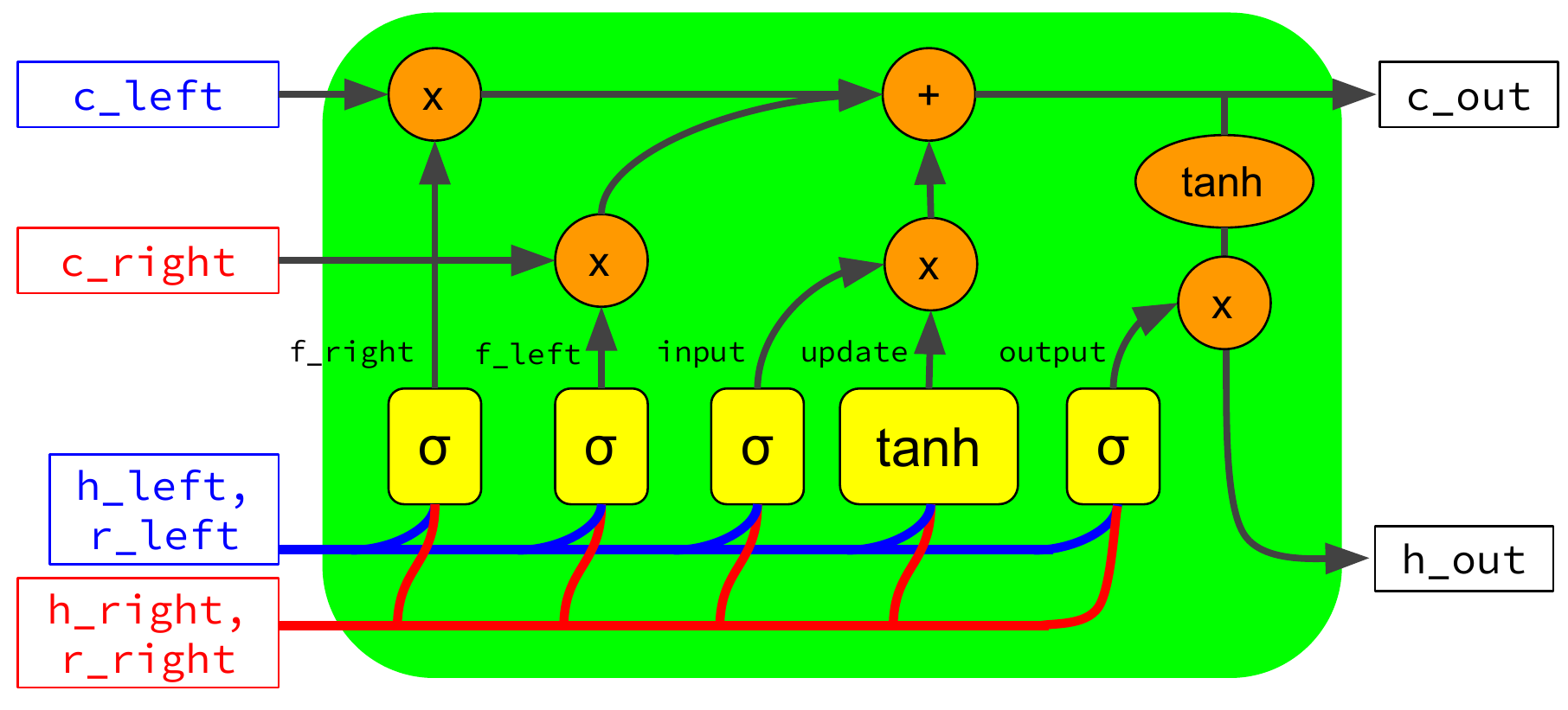}
    \vspace{-0.3in}
    \caption{Recursive LSTM architecture used in RST-Recursive adapted from \cite{socher-tree-lstm}.}
    \vspace{-0.2in}
    \label{fig:rst-recursive-lstm-architecture}
\end{figure}

Our work, however, is the first to develop and test a neural approach to leveraging RST discourse representations in coherence evaluation. Furthermore, \citet{feng-etal-2014-impact} only tested their proposal on the sentence permutation task, which involves ranking a sentence-permuted text against the original. As noted by \citet{lai-grammerly}, this is not an accurate proxy for realistic coherence evaluation. We evaluate our method on their more realistic Grammarly Corpus Of Discourse Coherence (GCDC), where the model needs to classify a naturally produced text into one of three levels of coherence. Our contributions involve: \textbf{(1)} RST-Recursive, an RST-based neural tree-recursive method for coherence evaluation that achieves 2\% below the state of the art performance on the GCDC while having 62\% fewer parameters. \textbf{(2)} When ensembled with the current state of the art, namely Parseq \cite{lai-grammerly}, we achieve a notable improvement over the plain ParSeq model. \textbf{(3)} We demonstrate the usefulness of silver-standard RST features in coherence classification, and establish our results as a lower-bound for performance improvements to be gained using RST features.

\section{Related Work}

\subsection{Coherence Evaluation of Text}

Centering Theory \cite{grosz-centering-theory} states that subsequent sentences in coherent texts are likely to continue to focus on the same entities (i.e., subjects, objects, etc.) as within the previous sentences. Building on top of this, \citet{Barzilay-Entity-Grid} were the first to propose the Entity-Grid model that constructs a two-dimensional array $G_{n,m}$ for a text of $n$ sentences and $m$ entities, which are used to estimate transition probabilities for entity occurrence patterns. More recently, \citet{elsner-charniak-2011-extending} extended Entity-Grid using entity-specific features, while \citet{tien-nguyen-joty-2017-neural} used a Convolutional Neural Network (CNN) on top of Entity-Grid to learn more hierarchical patterns. 

On the other hand, feature-free deep neural techniques have dominated recent research. \citeauthor{li-jurafsky-2017-neural} \shortcite{li-jurafsky-2017-neural} applied Recurrent Neural Networks (RNNs) to model the coherent generation of the next sentence given the current sentence and vice-versa. \citeauthor{mesgar-strube-2018-neural} \shortcite{mesgar-strube-2018-neural} constructed a local coherence model that encodes patterns of changes on how adjacent sentences within the text are semantically related. Recently, \citet{moon-etal-2019-unified} used a multi-component model to capture both local and global coherence perturbations. \citeauthor{lai-grammerly} \shortcite{lai-grammerly} developed a hierarchical neural architecture named ParSeq with three stacked LSTM Networks, designed to encode the coherence at sentence, paragraph and document levels.

\subsection{Rhetorical Structure Theory (RST)}

RST describes the structure of a text in the following way: first, the text is segmented into elementary discourse units (EDUs), which describe spans of text constituting clauses or clause-like units \cite{mann-RST-theory}. Second, the EDUs are recursively structured into a tree hierarchy where each node defines an RST relation between the constituting sub-trees. The sub-tree with the central purpose is called the \textit{nucleus}, and the one bearing secondary intent is called the \textit{satellite} while a connective discourse relation is assigned to both. An example of a “nucleus-satellite" relation pairing is presented in Figure \ref{fig:rst-recursive-overview} where a claim is followed by the evidence for the claim; RST posits an “Evidence” relation between these two spans with the left sub-tree being the “nucleus" and the right sub-tree as “satellite". 

\section{Method}

\subsection{RST-Recursive}
\label{RST-Recursive}

\begin{figure}[t]
    \centering
    \includegraphics[width=3.05in]{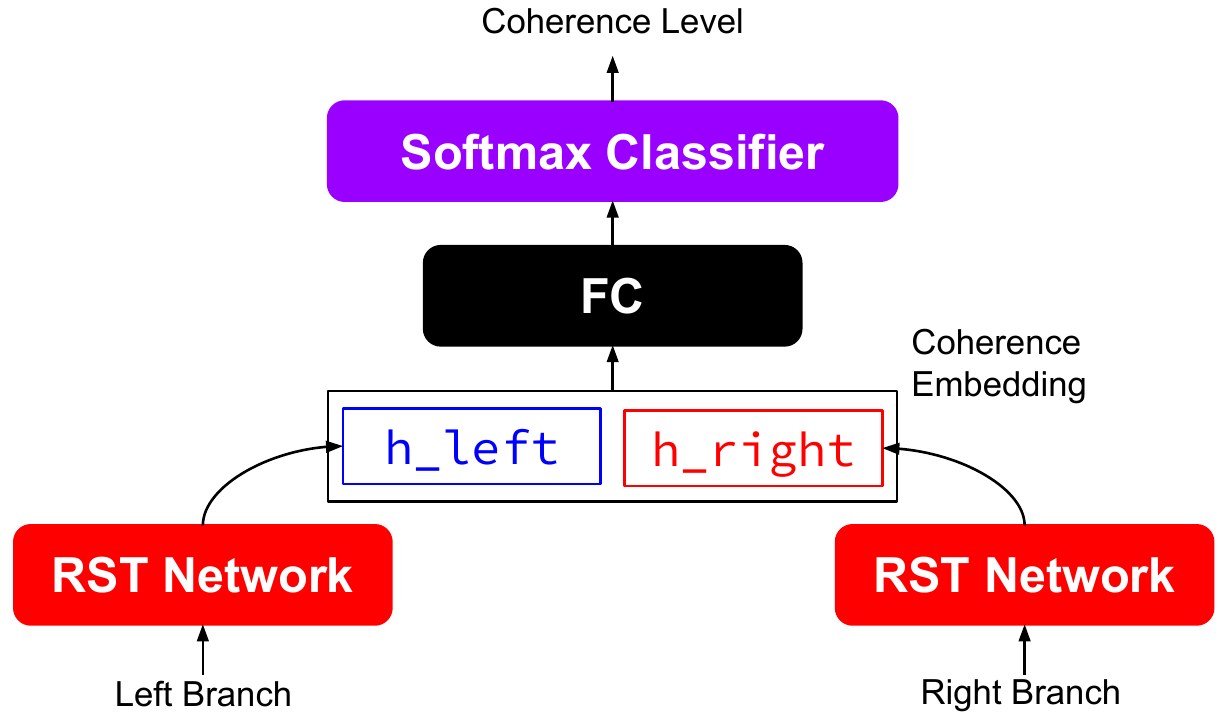}
    \vspace{-0.3in}
    \caption{Overview of the classification layer in RST-Recursive; At the root of the RST tree, children's hidden states are concatenated to form the document representation $\mathbf{d} = [\mathbf{h}_{l}, \mathbf{h}_{r}]$ which is then transformed into a 3-dimensional vector of Softmax probabilities.}
    \vspace{-0.2in}
    \label{fig:rst-recursive-final-layer}
\end{figure}

We parse silver-standard RST trees for documents using the CODRA \cite{joty-codra} RST parser, which we then employ as input to our recursive neural model, RST-Recursive. The overall procedure for RST-Recursive is shown in Figure \ref{fig:rst-recursive-overview}. Given a document of $n$ EDUs $\mathcal{E}_{1:n}$ with each EDU $\mathcal{E}_i$ represented as a list of GloVe embeddings \cite{pennington-etal-2014-glove}, we use an LSTM to process each $\mathcal{E}_i$, using the final hidden state as the EDU embedding $\mathbf{e}_i=\text{LSTM}(\mathcal{E}_i)$ for each leaf $i$ of the document's RST tree. Afterwards, we apply a recursive LSTM architecture (Figure \ref{fig:rst-recursive-lstm-architecture}) that traverses the RST tree bottom-up. At each node $s$, we use the children's sub-tree embeddings $[\mathbf{h}_{l}, \mathbf{c}_{l}, \mathbf{r}_{l}]$ and $[\mathbf{h}_{r}, \mathbf{c}_{r}, \mathbf{r}_{r}]$ to form the node's sub-tree embedding:
\begin{multline}
    [\mathbf{h}_s, \mathbf{c}_s] = \text{TreeLSTM}([\mathbf{h}_{l}, \mathbf{c}_{l}, \mathbf{r}_{l}], [\mathbf{h}_{r}, \mathbf{c}_{r}, \mathbf{r}_{r}])
    \label{main}
\end{multline}
where $\mathbf{h}_l$/$\mathbf{c}_l$ and $\mathbf{h}_{r}$/$\mathbf{c}_{r}$ are the LSTM hidden and cell states from the left and right sub-trees respectively. The relation embeddings of the children sub-trees, $\mathbf{r}_{l}$ and $\mathbf{r}_{r}$, are learned vector embeddings for each of the 31 pre-defined relation labels in the form of “[relation]\_[nucleus/satellite]" (e.g., “Evidence\_Satellite" for the last EDU in Figure \ref{fig:rst-recursive-overview}). At the root of the tree, the output hidden states from both children are concatenated into a single document embedding $\mathbf{d} = [\mathbf{h}_{l}, \mathbf{h}_{r}]$. As shown in Figure \ref{fig:rst-recursive-final-layer}, a fully connected layer is applied to this representation before using a Softmax function to obtain the coherence class probabilities.

\subsection{Ensemble: ParSeq + RST-Recursive}
To evaluate if the addition of silver-standard RST features to existing methods can improve coherence evaluation, we ensemble RST-Recursive with the current state of the art coherence classifier: ParSeq. 

\begin{figure}[t]
  \centering{\includegraphics[width=3.0in]{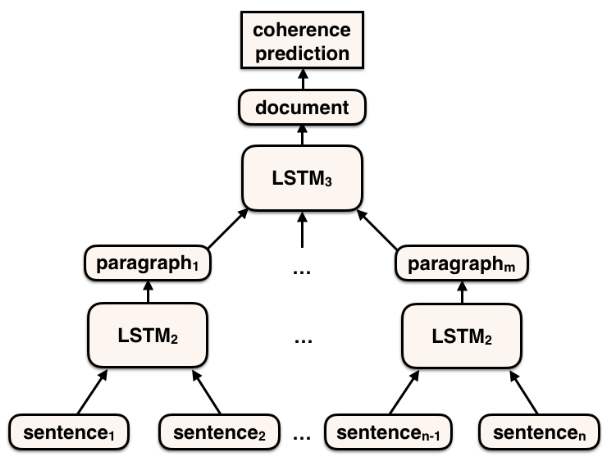}}
  \vspace{-0.35in}
\caption{The architectural overview of ParSeq; an illustration of ParSeq's structure, taken directly from the original paper \cite{lai-grammerly}.}
\label{grammarly-architecture}
\vspace{-0.1in}
\end{figure}

A deep learned non-linguistic classifier, ParSeq employs three layers of LSTMs that intend to capture coherence at different granularities. An overview of the ParSeq architecture is presented in Figure \ref{grammarly-architecture}. First, $\text{LSTM}_1$ (not shown) produces a single sentence embedding for each sentence in the text. Next, $\text{LSTM}_2$ generates paragraph embeddings using the corresponding sentence embeddings from $\text{LSTM}_1$. Finally, $\text{LSTM}_3$ reads the paragraph embeddings, generating the final document embedding, which is passed to a fully connected layer to produce Softmax label probabilities. 

In this augmented variation of our model, we operate ParSeq on the document independently until a document level embedding $\mathbf{d}_{p}$ is obtained at the highest-level LSTM. This document embedding is then concatenated to the RST-Recursive coherence embedding $\mathbf{d} = [\mathbf{h}_{l}, \mathbf{h}_{r}, \mathbf{d}_{parseq}]$ in Figure \ref{fig:rst-recursive-final-layer} to produce class probabilities. Note that in this ensemble variation, we initialize tree leaves $\mathbf{e}_{1:n}$ with zero-vectors as opposed to EDU embeddings since ParSeq is sufficiently capable of capturing semantic information on its own, and early experiments using 5-fold cross-validation on the training set revealed model overfitting when training with EDU embeddings simultaneously.

\begin{table*}[t]
\vskip 0.15in
\centering
\tabcolsep=0.13cm
\begin{sc}
 \begin{tabular}{l|cccc|cccccccccccc} 
 \toprule
 Model & T & NS & R & E & Clinton & Enron & Yahoo & Yelp & Average \\
 \midrule
 Majority & {} & {} & {} & {} & 55.33 & 44.39 & 38.02 & 54.82 & 48.14 \\
 \midrule
 RST-Rec & \checkmark & {} & {} & {} & 55.33$\pm$0.00 & 44.39$\pm$0.00 & 38.02$\pm$0.00 & 54.82$\pm$0.00 & 48.14$\pm$0.00 \\
 RST-Rec & \checkmark & \checkmark & {} & {} & 53.74$\pm$0.14 & 44.67$\pm$0.07 & 44.61$\pm$0.09 & 53.76$\pm$0.11 & 49.20$\pm$0.07 \\
 RST-Rec & \checkmark & \checkmark & \checkmark & {} & 54.07$\pm$0.10 & 43.99$\pm$0.07 & 49.39$\pm$0.10 & 54.39$\pm$0.12 & 50.46$\pm$0.05\\
 RST-Rec & \checkmark & \checkmark & \checkmark & \checkmark & 55.70$\pm$0.08 & 53.86$\pm$0.11 & 50.92$\pm$0.13 & 51.70$\pm$0.16 & 53.04$\pm$0.09 \\
 \midrule
 ParSeq & {} & {} & {} & {} & \textbf{61.05$\pm$0.13} & \textbf{54.23$\pm$0.10} & 53.29$\pm$0.14 & 51.76$\pm$0.21 & 55.09$\pm$0.09 \\
 \midrule
 Ensemble & \checkmark & {} & {} & * & \textbf{61.12$\pm$0.13} & \textbf{54.20$\pm$0.12} & 52.87$\pm$0.16 & 51.52$\pm$0.22 & 54.93$\pm$0.10 \\
 Ensemble & \checkmark & \checkmark & {} & * & 60.82$\pm$0.13 & 54.01$\pm$0.10 & 52.92$\pm$0.15 & 51.63$\pm$0.24 & 54.85$\pm$0.10 \\
 Ensemble & \checkmark & \checkmark & \checkmark & * & \textbf{61.17$\pm$0.12} & 53.99$\pm$0.10 & \textbf{53.99$\pm$0.14} & \textbf{52.40$\pm$0.21} & \textbf{55.39$\pm$0.09} \\
 \bottomrule
\end{tabular}
\end{sc}
\vspace{-0.05in}
\caption{Overall and sub-dataset specific coherence classification accuracy on the GCDC dataset. Error boundaries describe 95\% confidence intervals. Values in bold describe statistically significant state of the art performance. * indicates availability of EDU-level semantic information through the ensembling with ParSeq.}
\vspace{-0.2in}
\label{table-results-accuracy}
\end{table*}

\begin{table*}[t]
\vskip 0.15in
\centering
\tabcolsep=0.13cm
\begin{sc}
 \begin{tabular}{l|cccc|cccccccccccc} 
 \toprule
 Model & T & NS & R & E & Clinton & Enron & Yahoo & Yelp & Average \\
 \midrule
 Majority & {} & {} & {} & {} & 39.42 & 27.29 & 20.95 & 38.82 & 31.62 \\
 \midrule
 RST-Rec & \checkmark & {} & {} & {} & 39.42$\pm$0.00 & 27.29$\pm$0.00 & 20.95$\pm$0.00 & 38.82$\pm$0.00 & 31.62$\pm$0.00 \\
 RST-Rec & \checkmark & \checkmark & {} & {} & 39.20$\pm$0.03 & 30.81$\pm$0.16 & 35.67$\pm$0.18 & 39.93$\pm$0.08 & 36.40$\pm$0.09 \\
 RST-Rec & \checkmark & \checkmark & \checkmark & {} & 41.08$\pm$0.07 & 31.21$\pm$0.13 & 41.97$\pm$0.14 & 42.27$\pm$0.09 & 39.13$\pm$0.08 \\
 RST-Rec & \checkmark & \checkmark & \checkmark & \checkmark & 45.90$\pm$0.12 & 44.33$\pm$0.16 & 43.85$\pm$0.18 & 43.13$\pm$0.10 & 44.30$\pm$0.08\\
 \midrule
 ParSeq & {} & {} & {} & {} & \textbf{52.12$\pm$0.21} & \textbf{44.90$\pm$0.15} & 46.22$\pm$0.18 & 43.36$\pm$0.09 & 46.65$\pm$0.10 \\
 \midrule
 Ensemble & \checkmark & {} & {} & * & \textbf{52.35$\pm$0.22} & \textbf{44.92$\pm$0.16} & 45.48$\pm$0.22 & 43.70$\pm$0.11 & 46.61$\pm$0.11 \\
 Ensemble & \checkmark & \checkmark & {} & * & 51.90$\pm$0.22 & 44.76$\pm$0.14 & 45.48$\pm$0.22 & \textbf{43.83$\pm$0.13} & 46.49$\pm$0.10 \\
 Ensemble & \checkmark & \checkmark & \checkmark & * & \textbf{52.42$\pm$0.19} & 44.69$\pm$0.15 & \textbf{46.88$\pm$0.17} & \textbf{43.94$\pm$0.09} & \textbf{46.98$\pm$0.09} \\
 \bottomrule
\end{tabular}
\end{sc}
\vspace{-0.05in}
\caption{Overall and sub-dataset specific coherence classification F1 scores on the GCDC dataset. Error boundaries describe 95\% confidence intervals. Values in bold describe statistically significant state of the art performance. F1 scores are calculated by macro-averaging the corresponding class-wise F1 scores. * indicates availability of EDU-level semantic information through the ensembling with ParSeq.}
\vspace{-0.15in}
\label{table-results-f1}
\end{table*}

\begin{figure*}
    \centering 
    \includegraphics[width=6.35in]{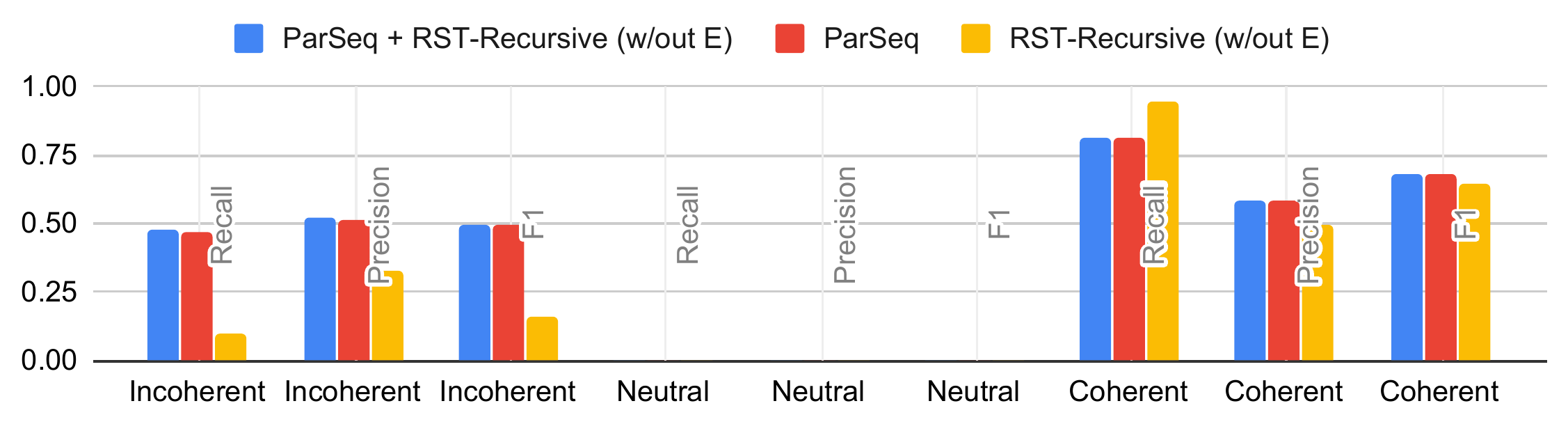}
    \vspace{-0.35in}
    \caption{Comparison of Recall, Precision and F1 on overall classification of each coherence level.}
    \vspace{-0.2in}
    \label{fig:class_comparison}
\end{figure*}

\section{Experiments}

\subsection{Dataset}

We evaluate RST-Recursive and Ensemble on the GCDC dataset \cite{lai-grammerly}. This dataset consists of 4 separate sub-datasets: Clinton emails, Enron emails, Yahoo answers, and Yelp reviews, each containing 1000 documents for training and 200 documents for testing. Each document is assigned a discrete coherence label of incoherent (1), neutral (2), and coherent (3). 

We parse RST trees for each example within the GCDC dataset using CODRA \cite{joty-codra}. Due to CODRA's imperfect parsing of documents, RST trees could not be obtained for approximately 1.5\%-2\% of the documents, which were then excluded from the study. In addition, we re-evaluated ParSeq on only the RST-parsed portion of documents to assure consistent comparability of results. For more details, see Appendix \ref{appendix:dataset}/\ref{appendix:codra}.  Our code and dataset can be accessed below\footnote{https://github.com/grig-guz/coherence-rst}, and the access to the original GCDC corpus can be obtained here\footnote{https://github.com/aylai/GCDC-corpus}. We can share RST-parsings of GCDC examples with interested readers upon request once access to the GCDC dataset has also been obtained.

\subsection{Training}

We train all models with hyperparameter settings consistent with that of ParSeq reported by \cite{lai-grammerly}. Specifically, we use a learning rate of 0.0001, hidden size of 100, relation embedding size of 50, and 300-dimensional pre-trained GloVe embeddings \cite{pennington-etal-2014-glove}. We train with the Adam optimizer \cite{Adam} for 2 epochs. For every model/variation, the reported results represent the corresponding accuracies and F1 scores averaged over 1000 independent runs, each initialized with a different random seed.

\subsection{RST-Recursive's Performance}
Our full model incorporates the RST Tree (T) structure, nucleus/satellite properties (nuclearity) of sub-trees (NS), RST specific connective relations (R), and EDU embeddings at leaves of the RST tree (E), as previously described in \ref{RST-Recursive}. Here, (T) defines the tree traversal operation and (NS) and (R) are learned vector embeddings for nuclearity and relations. We examine three ablations, each removing one of (NS), (R) and (E) from the model. 

The results are provided in Tables \ref{table-results-accuracy} and \ref{table-results-f1}. As shown, the complete model is able to achieve a competitive overall accuracy and F1 at 53.04\% and 44.30\% respectively, which is close to the state of the art. Although this lags behind ParSeq by a noticeable 2\% 
margin, RST-Recursive is able to achieve this performance with 62\% fewer parameters (1,230k vs. 3,241k), demonstrating the usefulness of linguistically-motivated features. Removing EDU embeddings reduces accuracy and F1 scores to 50.46\% and 39.13\%. This is still significantly better than the majority class baseline, signifying that even without any semantic information about the text and its contents, it is still possible to evaluate coherence using just the silver-standard RST features of the text. Removing RST relations and nuclearity, however, decreases performance substantially, dropping to the majority class level. This indicates that an RST tree structure alone (of the quality delivered by silver-standard parsers) is not sufficient to classify coherence. It must also be noted that since we employ silver-standard RST parsing as performed by CODRA \cite{joty-codra}, the reported results act as a lower bound which we would expect to improve as parsing quality increases.

\subsection{Ensemble's Performance}
We examine three variations of the Ensemble. The full model augments ParSeq with the text's RST tree, relations and nuclearity. This model is able to achieve the new state of the art performance, at 55.39\% accuracy and 46.98\% F1. Using final layer concatenation for ensembling is widely applicable to many other neural methods, and serves as a lower bound for the accuracy/F1 boost to be appreciated by incorporating RST features into the model. Removing the RST relations and/or nuclearity information completely eliminates the performance gain, which shows that the RST tree on its own is not sufficient as an RST source of information for distinguishing coherence, even when ensembled with ParSeq.

\subsection{Classification Trends}
As demonstrated in Figure \ref{fig:class_comparison}, coherence classifiers have difficulty predicting the neutral class (2), experiencing modal collapse towards the extreme ends in the best performing models. Early experiments using alternative objective functions such as the Ordinal Loss or Mean Squared Error resulted in a similar modal collapse or poor overall performance. We leave further exploration of this problem to future research. Furthermore, RST-Recursive shows a notably stronger recall on the coherent class (3) as compared to ParSeq. On the other hand, ParSeq has a higher recall/precision on class (1) and slightly higher precision on class (3). The Ensemble method, however, is able to take the best of both, achieving better recall, precision and F1 on both the incoherent and coherent classes as compared to ParSeq.

\section{Conclusions and Future Work}
In this paper, we explore the usefulness of silver-standard parsed RST features in neural coherence classification. We propose two new methods, RST-Recursive and Ensemble. The former achieves reasonably good performance, only 2\% short of state of the art, while more robust with 62\% fewer parameters. The latter demonstrates the added advantage of RST features in improving classification accuracy of the existing state of the art methods by setting new state of the art performance with a modest but promising margin. This signifies that the document's rhetorical structure is an important aspect of its perceived clarity. Naturally, this improvement in performance is bounded by the quality of parsed RST features and could increase as better discourse parsers are developed. 

In the future, exploring other RST-based architectures for coherence classification, as well as better RST ensemble schemes and improving RST parsing can be avenues of potentially fruitful research. Additional research on multipronged approaches that draw from Centering Theory, RST and deep learning all together can also be of value.

\bibliography{anthology,aacl-ijcnlp2020}
\bibliographystyle{acl_natbib}

\newpage
\begin{appendices}

\begin{table}[!b]
\centering
\small
 \begin{tabular}{p{2.8in}} 
 \toprule
 \textbf{Coherence} / Example \\
 \midrule
 \footnotesize \textbf{Incoherent (1)} \\
 For good Froyo, you just got to love some MoJo, yea baby yea! Creamy goodness with half the guilt of ice cream, a spread of tasty toppings, this in the TMP in definitely the place to be! They have little cups for sampling to find your favorite flavor. Great prices and with a yelping good 25\% off discount just for "checking in" and half off Tuesdays with the FB word of the day, you just can't beat it! Perfect summer treat located in front of the TMP splash pad, you can soak up some sun and enjoy some fromazing yogurt in their outdoor sitting area! Go get you some  Mojo froyo! \\
 \midrule
 \footnotesize \textbf{Neutral (2)} \\ 
 So Spintastic gets 5 stars because it's about as good as it gets for a laundromat, me thinks.
 
 Came here bc the dryer at my place was busted and waiting on the repairman.  I found the people working the place extremely helpful.  It was my first time there and she walked me through the steps of how to get a card, which machines to use, where I could buy the soap... only thing she didn't do was fold my dried laundry! Heh.
 
 Will remember this place for the future in the event that I need to get my clothes washed and ready.  Free wi-fi and a soda machine is convenient.  Oh and if you have a balance left on your card, you can redeem the card and any remaining balance if you like. 
 
 dmo out \\
 \midrule
 \footnotesize \textbf{Coherent (3)} \\ 
 vet for almost 6 years.  He is kind, compassionate and very loving and gentle with my dogs.  All my dogs are shelter dogs and I am very picky about who cares for my animals.
 
 I walked in once with a dog I found running around the neighborhood and the staff could not find a chip so Dr. Besemer came out to help. He was busy but made time for me.  He looked over the dog and could not find a chip, he also did a quick check on the dog and said that he appeared healthy. He didn't charge me for his time.   This dog became my third adoped dog.  Dr. Besemer is the best and I highly recommend him if you are looking for a vet.  His staff is kind and compassionate. \\
 \bottomrule
\end{tabular}
\vspace{-0.05in}
\caption{Text examples of incoherent (class 1), neutral (class 2), and coherent (class 3) snippets from the Yelp subset of the GCDC dataset \cite{lai-grammerly}.}
\label{grammerly-examples}
\vspace{-0.1in}
\end{table}

\begin{table}[t!]
\centering
\begin{small}
 \begin{tabular}{c c c c c} 
 \toprule
 Parser & Structure & Nuclearity & Relation & Full \\
 \midrule
 CODRA & 82.6 & 68.3 & 55.8 & 55.4 \\
 Human & 88.3 & 77.3 & 65.4 & 64.7 \\
 \bottomrule
\end{tabular}
\end{small}
\vspace{-0.05in}
\caption{Micro-averaged F1 scores on the RST parsing of text by CODRA vs. Human Standard \cite{Morey2017HowMP}.}
\label{Table-Parser}
\vspace{-0.1in}
\end{table}

\begin{table*}[t!]
\vskip 0.15in
\begin{center}
\begin{small}
\begin{sc}
 \begin{tabular}{l|cccc|cccc} 
 \toprule
 {} & \multicolumn{4}{c}{Train} & \multicolumn{4}{c}{Test} \\
 {} & Clinton & Enron & Yahoo & Yelp & Clinton & Enron & Yahoo & Yelp \\
 \midrule
 Examples & 1000 & 1000 & 1000 & 1000 & 200 & 200 & 200 & 200 \\
 Pre-Fix RST-Trees & 667 & 710 & 940 & 950 & 136 & 142 & 188 & 190 \\
 Post-Fix RST-Trees & 985 & 976 & 986 & 999 & 199 & 195 & 192 & 197 \\
 \midrule
 Post-Fix Very coherent & 503 & 499 & 368 & 511 & 109 & 87 & 73 & 109 \\
 Post-Fix Medium coherent & 204 & 192 & 170 & 218 & 38 & 50 & 41 & 42 \\
 Post-Fix Incoherent & 277 & 289 & 442 & 270 & 50 & 59 & 78 & 47 \\
 \bottomrule
\end{tabular}
\end{sc}
\end{small}
\end{center}
\vspace{-0.1in}
\caption{Number of examples for which RST trees were successfully produced in each GCDC sub-dataset.}
\label{Table-Dataset}
\vspace{-0.1in}
\end{table*}

\section{Dataset Description}
\label{appendix:dataset}

For model evaluation, we use the recently released Grammarly Corpus for Discourse Coherence \cite{lai-grammerly}. GCDC consists of 4 sections - Clinton and Enron emails, as well as Yelp review and Yahoo answers, with 1000 training and 200 testing examples in each section. Each text is given a score from 1 (least coherent) to 3 (most coherent) by expert raters. GCDC's key advantage, compared to the ranking corpora used in the past \cite{prasad-penntb}, is that all the datapoints are human-labelled and not artificially permuted. Examples from the dataset are provided in Table \ref{grammerly-examples}. When assigning the ranking to each text, the experts received the following instructions \cite{lai-grammerly}:

\textit{ A text that is highly coherent (score 3) is easy to understand and easy to read. This usually means the text is well-organized, logically structured, and presents only information that supports the main idea. On the other hand, a text with low coherence (score 1) is difficult to understand. This may be because the text is not well organized, contains unrelated information that distracts from the main idea, or lacks transitions to connect the ideas in the text. Try to ignore the effects of grammar or spelling errors when assigning a coherence rating.}

We generated a discourse tree for each text in the GCDC dataset, utilizing the available CODRA discourse parser \cite{joty-codra}. Early iterations resulted in up to 30\% unsuccessful parsing rate on some sub-datasets. As a result, a punctuation fixing script was developed to fix minor punctuation problems without changing the text's structure or coherence. Post-fixing results lowered this RST parsing failure rate to reasonable margins in the 1\% to 3\% region (see Table \ref{Table-Dataset}). Note that all examples for which RST parsing was not successfully performed were excluded in our experiments. All baselines were re-evaluated using the RST-parsed set of examples.

\section{CODRA Quality}
\label{appendix:codra}
While partial parsing of the dataset (see Appendix \ref{appendix:dataset}) allows us to evaluate the accuracy of our models, it must be emphasized that as with the goal of this paper, we've used silver-standard RST parsing which lags well behind the human gold-standard. As shown in Table \ref{Table-Parser}, CODRA is far from reaching human-level accuracy in RST parsing. Additionally, since it was trained on RST-DT \cite{carlson-rst}, it lacks out-of-domain adaptability, which becomes a bottle-neck in achieving substantial performance boost on badly structured domains of text such Yelp review. We again re-iterate the importance of RST parsing for RST-based coherence evaluation, and motivate future work in this area. We believe that improvements in RST parsing will result in better accuracy for both future and existing RST-based coherence evaluation methods.

\end{appendices}

\end{document}